# Towards Sustainable Artificial Intelligence: An Overview of Environmental Protection Uses and Issues

**Arnault Pachot**
OpenStudio
apachot@openstudio.fr

**Céline Patissier**
OpenStudio
cpatissier@openstudio.fr

Artificial Intelligence (AI) is used to create more sustainable production methods and model climate change, making it a valuable tool in the fight against environmental degradation. This paper describes the paradox of an energy-consuming technology serving the ecological challenges of tomorrow. The study provides an overview of the sectors that use AI-based solutions for environmental protection. It draws on numerous examples from AI for Green players to present use cases and concrete examples. In the second part of the study, the negative impacts of AI on the environment and the emerging technological solutions to support Green AI are examined. It is also shown that the research on less energy-consuming AI is motivated more by cost and energy autonomy constraints than by environmental considerations. This leads to a rebound effect that favors an increase in the complexity of models. Finally, the need to integrate environmental indicators into algorithms is discussed. The environmental dimension is part of the broader ethical problem of AI, and addressing it is crucial for ensuring the sustainability of AI in the long term.

## 1 INTRODUCTION

The widespread adoption of Deep Learning in a variety of fields has led to a surge of interest in AI technology and has sparked significant investments in AI startups. Deep Learning has been applied to tasks such as image recognition, language processing, strategy games, and even music generation, demonstrating its potential for further development and expansion in the coming years.

However, as the complexity and performance of AI models continue to improve, it is important to consider the environmental impact of these models. AI models can require large amounts of computational power and energy, which can have negative consequences for the environment. In our study, we will explore the use of AI for environmental applications and will examine the environmental issues associated with AI models. We will show that while efficiency may increase the complexity of AI models, it does not necessarily reduce the overall environmental impact of AI. We will conclude by discussing the need for regulations and the qualification.

## 2 AI FOR GREEN

AI has many applications in environmental conservation, including in the management of smart cities, energy, agriculture, natural disaster prediction and adaptation to climate change, ecosystem preservation, mobility, and the economy.

### 2.1 Smart Cities

The rise of smart cities is a growing trend in urban development [34][41]. Smart cities use technology, data, and intelligent systems to improve the quality of life for their citizens. These cities are designed to be more efficient, sustainable, and livable, and they use a range of technologies, such as the internet of things (IoT), artificial intelligence (AI), and big data, to achieve these goals. Smart cities are often seen as the future of urban development, and many cities around the world are implementing smart city initiatives to improve their infrastructure and services.

AI is being used to manage traffic lights in a more efficient manner, reducing the amount of time vehicles spend stopped and therefore reducing emissions by 20% [25][65]. In China, AI is used to anticipate air pollution and take preventive measures before dangerous levels are reached [26].

One way to solve the challenges of street cleanliness and waste management in municipalities is to use AI to optimize resources and improve efficiency [33]. AI solutions such as the intelligent waste containers developed by Bin-e [27] and the AI-powered sorting robot are emerging to help manage waste more responsibly [28].

By using AI, it is possible to anticipate a city's need for energy resources and limit unnecessary expenditure. AI can also be used to reduce the effects of climate change or natural disasters by making urban planning more intelligent. For example, the City of Los Angeles has launched the Tree Canopy Lab program, which uses AI to map the city and recommend where trees should be planted to prevent heat [35]. The development of "urban dashboards" with real-time data on all environmental parameters, such as water and energy consumption, traffic pollution, and weather conditions, could help cities become more environmentally responsible and improve the quality of life of their inhabitants.

## 2.2 Energy

AI has the potential to play a significant role in the transition to renewable energy sources. By analyzing large amounts of data, AI can help optimize the performance of wind farms and other renewable energy systems, improving their efficiency and reducing their environmental impact. Additionally, AI can be used to anticipate energy demand and identify ways to reduce energy consumption, further contributing to the transition to a more sustainable energy system.

One way to save energy is to promote renewable energy as a substitute for fossil fuels. But it is still necessary for renewable energies to become efficient enough to be used on a massive scale. In this case, AI is proving to be very useful. Thanks to the intervention of AI, it is possible to significantly improve the performance of wind farms by considering meteorological data. AI is used to correlate the speed of each propeller with the direction and power of the wind, which allows for optimization of electricity production from all the wind turbines. The European Centre for Medium-Range Weather Forecasts (ECMWF) led the Energy-efficient Scalable Algorithms for Weather Prediction at Exascale (ESCAPE) project, which aimed to develop a sustainable strategy for evolving weather and climate prediction models for next-generation computing technologies. The project involved leading European regional forecasting consortia, university research, experienced high-performance computing centers, and hardware vendors [36].

Some companies, like Google and Huawei, have already implemented AI solutions to control energy consumption in their data centers. Google has reduced its energy consumption by 40% using AI to analyze the times of day when people do energy-consuming searches and optimize the cooling of its data centers [23]. Huawei has improved the energy efficiency of its data centers by using AI to identify and address factors that contribute to increased energy consumption, as well as to predict the future energy efficiency of its data centers [22]. In addition, Microsoft has partnered with Vattenfall to develop a smart grid management solution that optimizes the production of renewable energy based on demand [14].

The application of AI technologies in smart buildings through building management systems and demand response programs has the potential to improve urban energy efficiency [38][39]. Governments are using AI to prioritize the energy renovation of its public buildings [40].

Energy efficiency can also be applied to digital technologies, particularly in the visualization and display of photos and videos online. Since 2017 Google has also been using AI to compress its images and reduce bandwidth consumption[9]. JPEG has also launched a program to find ways to reduce the size of its photo formats without losing quality using AI[7]. Other companies, like Netflix, have used AI to optimize the consumption of their videos, allowing them to halve their bandwidth consumption without losing broadcast quality[8].

## 2.3 A connected and sustainable agriculture

AI has the potential to greatly improve the sustainability and efficiency of agriculture [43][44]. By collecting and analyzing data from various sources, AI algorithms can help farmers make informed decisions about crop production, water usage, and pest management. This can help reduce the use of pesticides and other harmful chemicals, as well as increase crop yields and decrease water waste [31][32]. Improper irrigation and soil management can lead to crop loss and reduced crop quality. Examples of such systems include those that evaluate the design and performance of micro irrigation systems, recommend crops based on land suitability maps, and estimate soil moisture content [43]. These AI-based systems have been found to be effective in improving irrigation and soil management practices. The agricultural industry has the potential to greatly benefit the environment and improve the sustainability of food production.

AI has been used to predict crop diseases and recommend control measures. Hybrid systems that integrate image processing with AI have also been developed. For example, "Dr. Wheat" is a web-based expert system that uses AI to diagnose wheat diseases [42].

## 2.4 *Anticipate natural disasters and adapt to climate change*

The use of AI in anticipating natural disasters and adapting to climate change has the potential to greatly improve the way we respond to and prepare for such events [45][46]. By analyzing large amounts of data and making predictions based on that data, AI can help us to better understand the likelihood of natural disasters occurring and what areas may be at risk. This information can then be used to develop plans and strategies for mitigating the impact of these disasters and reducing the potential for harm to people and property. Additionally, AI can help us to better understand and adapt to climate change by analyzing data on weather patterns, climate trends, and other factors that may be contributing to changes in our environment. This information can be used to develop strategies for adapting to these changes and minimizing their impact.

AI can help farmers improve weather forecasts and anticipate extreme weather events, which can be invaluable in protecting crops. By using its calculation power and the ability to process large amounts of meteorological data, AI can provide farmers with valuable information to help them prepare for and adapt to unpredictable weather. This can help farmers avoid losing entire harvests due to unexpected weather events.

Natural disasters can have significant impacts on global trade and agriculture. Tropical storms, known as typhoons, hurricanes, or cyclones, seem to be occurring more frequently and with greater strength, causing significant economic damage to agricultural products, shipping efficiency, property, and airline rerouting. Predicting the direction and likelihood of land impacts of tropical storms would greatly increase the chances of insuring against potential damage [54]. Forecast-based financing is a financial mechanism that enables humanitarian actions in anticipation of floods by releasing pre-allocated funds based on the exceedance of flood forecast thresholds [47][55].

## 2.5 Preserve Earth's ecosystem

AI has the potential to play a crucial role in preserving wildlife and flora. Through the analysis of large amounts of data, AI can help researchers better understand the effects of climate change on biodiversity and make predictions about which species are most at risk. AI can also be used to identify and monitor individual animals and plants, providing valuable information for conservation efforts. Additionally, AI can help detect and prevent illegal activities such as poaching and deforestation. By using AI to better understand and protect ecosystems, we can work towards preserving the rich diversity of life on our planet.

AI has many potential applications in the field of environmental conservation. One such example is the use of AI in the monitoring of the Chesapeake Bay in the United States. Using ultra-fine image analysis, the area can be mapped more accurately, making it easier to monitor the bay and its biodiversity [56]. Another example is the partnership between Microsoft and the Nature Conservancy to map all ocean species using AI. This will help determine which areas can be used by humans without endangering the ecosystem [18]. Other initiatives using AI include the Ocean Cleanup project, which uses robots to clean up water on a large scale [15].

AI is being used to monitor terrestrial biodiversity and to help identify endangered species. The University of Southern California has set up a project called "Protection Assistant for Wildlife Security" (PAWS) that uses AI to predict where and when poachers are likely to strike [16]. This information can be used to arrest poachers and prevent the extinction of protected species. AI is also being used to help restore nature in areas damaged by human activity. In Massachusetts, an area destroyed by cranberry production has been rehabilitated and MIT Media Lab researchers are using microphones and AI to listen to interactions between species and determine the effectiveness of the restoration efforts [17].

## 2.6 **Autonomous transport and sustainable mobility**

AI can limit transport-related pollution by promoting fuel-efficient driving and optimizing engines to be more efficient. Companies in the automotive industry are heading towards innovation in autonomous and sustainable transport. Manufacturers are working on the development of shared, smart, and ecological transport. This is the ambition of the French company Transdev, a mobility specialist, which has partnered with ZF (a German automotive supplier) and e.Go (an electric car manufacturer) to develop its ecological and autonomous shuttle:

the e.Go Mover[6]. With a capacity of 15 people, these electric shuttles aim to complement existing urban transport networks. Other manufacturers have also entered the innovation race to come up with their own shared transport solution, following the example of Transdev, which shows that the trend is heading towards autonomous mobility but within communities. It is also conceivable that an attractive public transport network, both ecological and personalized, could trigger the abandonment of individual cars, at least for daily commutes.

### 2.7 Local and sustainable economy

Open data and artificial intelligence can facilitate the shift to more sustainable and environmentally friendly production models. Recent health and geopolitical crises have highlighted the fragility of Western industrial policies. The current awareness opens the way to a reindustrialization model that combines ecology, resilience, social commitment, and economic performance. New dimensions are being embraced, such as environmental and social impact, securing essential goods, supply chain robustness, and distributed manufacturing, promoting local production. The recent release of economic and industrial data, combined with the emergence of AI, offers new perspectives for building a digital twin of countries' productive systems that combine both macroeconomic data and real observations of each company's activity [51]. This modeling allows the identification of industrial know-how, value chains, and potential synergies between companies to build a sustainable industrial symbiosis [50][53]. AI could be used to identify synergistic pairings of one company's waste output with another company's input, facilitated by collaboration between companies through resource and information sharing [52].

## 3 EMERGENCE OF GREEN AI AND REBOUND EFFECTS

### 3.1 An energy consuming technology

In recent years, concerns about global warming and the depletion of resources have led to increased awareness of the environmental impact of the digital world. This has become a topic of public debate in many countries. After years of denial, the environmental impact of digital technology is now being recognized as a significant issue in research, including the environmental impact of terminal manufacturing, the energy required to use digital services, and the end-of-life analysis of equipment. Deep Learning is no exception to these concerns. In fact, due to the large amount of data and computational power required for Deep Learning, it has a significant impact on the environment. These increasing demands for computational power also contribute to the obsolescence of hardware and software.

Researchers measured the $CO_2$ emissions associated with the development of a natural language processing model, which generated the same amount of $CO_2$ as 5 cars over their lifetime, and the equivalent of 315 round trips by plane between New York and San Francisco [1]. This study is significant because it considered the 3200 learning iterations that were necessary to develop the final model. This work has contributed to a growing awareness of the environmental impact of AI, although it should be noted that the evaluation of AI methods still often focuses on precision and accuracy, without considering environmental factors such as the energy efficiency of the models.

Rohde et al. have detailed the different measures of energy consumption associated with tasks such as image classification, speech recognition, and strategy games [20]. These energy consumption levels are correlated with the complexity of the computations required, expressed in Peta-FLOPS/s-days, or 1015 floating point operations per second in one day [61]. The more complex the AI models, the higher the associated energy consumption, $CO_2$ emissions, and resource requirements.

### 3.2 The Importance of Ecodesign in AI Development

Deep Learning is a powerful approach to AI that has achieved impressive results in a variety of tasks, such as image recognition, natural language processing, and medical diagnosis. However, one of the challenges of Deep Learning is that it requires large amounts of training data to achieve good performance. In contrast, the human brain can learn from relatively few examples and can generalize to new situations. This suggests that there is still room for improvement in the mechanisms of learning used in Deep Learning, and that new approaches may be able to learn more efficiently from less data [11].

One potential direction for improving the efficiency of learning in Deep Learning models is to draw inspiration from the mechanisms of learning in the brain. For example, the brain can learn from a small number of examples by making use of prior knowledge and by using mechanisms such as attention and memory. These mechanisms could be incorporated into Deep Learning models to make them more efficient and more effective at learning from small amounts of data.

Another approach to improving the efficiency of learning in Deep Learning models is to develop more sophisticated optimization algorithms. Many Deep Learning models are trained using gradient-based optimization, which can be slow and require a large amount of data to converge to a good solution. New optimization algorithms, such as evolutionary algorithms or Bayesian optimization, may be able to find good solutions more quickly and with less data [62][63][64].

Overall, there is still much work to be done to improve the efficiency of learning in Deep Learning models. By drawing inspiration from the mechanisms of learning in the brain and by developing new optimization algorithms, researchers may be able to develop Deep Learning models that are more efficient and more effective at learning from small amounts of data.

### 3.3 Optimized Electronic Components for AI

Deep artificial neural networks use principles of the brain's information processing to make breakthroughs in machine learning in many problem domains. Neuromorphic computing aims to create chips inspired by the form and function of biological neural circuits, so they can process new knowledge, adapt, behave, and learn in real time at low power levels [58]. Intel claims that the new Loihi neuromorphic chip is 10,000 times more energy efficient than a CPU [10]. These chips are designed to handle the large amounts of data and complex calculations required by AI algorithms, making them more energy efficient and faster than traditional transistors.

### 3.4 Greening of Data Centers

Efforts to improve the efficiency of AI architectures must go hand in hand with a greening of the data center value chain[13]. The environmentally responsible hosting market is becoming more structured with the emergence of an energy performance indicator (called PUE) since 2007 by the consortium The Green Grid[12]. This indicator has been supplemented by the European DCEM (Data Center Energy Management) indicator, which also considers reused and renewable energy. By using renewable energy, cooling servers with natural resources or reusing the heat emitted, environmentally responsible data centers represent the most important lever in reducing the $CO_2$ emissions necessary for the operation of AI. For instance, we can cite the example of the Green Mountain data center in Norway, which, by cooling its servers with fjords and rivers, has been able to cut its energy costs by more than half [60].

### 3.5 The Rise of Hybrid AI

Hybrid AI is a type of artificial intelligence that combines multiple AI approaches or technologies in order to achieve more accurate, efficient, or flexible performance. By combining different AI methods, hybrid AI systems can leverage the strengths of each method to overcome the limitations of individual approaches.

The DesCartes program aims to develop a disruptive hybrid AI to serve the smart city and enable optimized decision-making in complex situations involving critical urban systems [59]. The program brings together experts in AI, engineering, data science, signal processing, formal methods, trusted AI, human-computer interaction, language processing, images, and the human sciences.

### 3.6 Continuous improvement of the energy efficiency of Deep Learning

Over the past few years, Deep Learning models have doubled their performance every 16 months [19][3][61]. This rapid progress in Deep Learning has been driven by a combination of factors, including increased computational power, larger and more diverse datasets, and more sophisticated algorithms and architectures.

One of the key drivers of this rapid progress is the availability of more powerful computational hardware, such as GPUs and TPUs, which are specifically designed to accelerate the training of Deep Learning models. These hardware platforms have allowed researchers to train larger and more complex Deep Learning models, which in turn have led to improved performance on a variety of tasks, such as image recognition, natural language processing, and medical diagnosis.

Another factor contributing to the rapid progress of Deep Learning is the availability of larger and more diverse datasets, which are used to train and evaluate Deep Learning models. These datasets have grown and complexity, and have become more representative of real-world data, allowing Deep Learning models to learn more robust and generalizable patterns.

Finally, the development of new algorithms and architectures for Deep Learning has also played a role in the rapid progress of Deep Learning. Researchers have proposed new ways of training and regularizing Deep Learning models, which have led to improved performance on a variety of tasks. These advances in Deep Learning algorithms and architectures have allowed researchers to build more powerful and effective Deep Learning models.

Overall, the rapid progress of Deep Learning over the past few years has been driven by a combination of increased computational power, larger and more diverse datasets, and more sophisticated algorithms and architectures.

### 3.7 *The Rebound Effect of improving energy efficiency in Artificial Intelligence*

Since the emergence of Deep Learning methods in 2012, AI has become a digital technology that requires more and more data. The increased computing power of computers has made it possible to increase the number of parameters and layers (for example, 530 billion parameters in the case of GPT3 "Megatron-Turing NLG" which required 4480 GPUs)[4][5]. Training a model with one trillion parameters would require 42,000 petaFLOPS-days, costing $19.2 million on Google's TPUs. One of the largest models published in 2020 used 600,000 times more computing power than the 2012 model that popularized Deep Learning. Since 2012 the growth of AI models has not followed Moore's law, with models doubling every 3.4 months [61].

## 4 CONCLUSION AND PERSPECTIVES

As the use of AI technologies continues to increase, the sustainability of these systems has become a major concern [49][57]. AI requires non-renewable natural resources, and its use must be managed responsibly. The scientific community must address the ecological issues of AI, just as it has addressed ethical and transparency issues [24]. Given the climate challenges, regulating AI will be necessary to evaluate the usefulness or futility of models to achieve the Sustainable Development Goals [29][30].

To move towards a more sustainable approach, Deep Learning and the large amounts of data and computation it requires must be replaced with less energy-consuming AI technologies. Projects that focus on optimizing energy use and transparency on the environmental impact of AI solutions, such as the Carbontracker, are crucial for achieving sustainability in AI [2]. The European Commission's proposed regulation of AI, the Artificial Intelligence Act (AI Act), seeks to create a uniform legal and regulatory framework for all AI in all sectors (except the military), and for all types of AI. As a product regulation, it does not grant rights to individuals but regulates the providers of AI systems and entities that use them for professional purposes. Unfortunately, this initiative does not consider the environmental impact of AI [48]. The "Sustainability Index for Artificial Intelligence" project aims to develop a set of sustainability criteria for AI-based systems and establish guidelines for sustainable AI development [21][20].

Combining AI and energy transition is a challenging but necessary task. By prioritizing sustainability, the AI community can ensure that AI continues to be a valuable tool in a responsible and ethical way.